\documentclass[conference]{IEEEtran}
\IEEEoverridecommandlockouts
\usepackage{cite}
\usepackage{amsmath,amssymb,amsfonts}
\usepackage{algorithmic}

\usepackage{todonotes,soul}
\usepackage{textcomp}
\usepackage{xcolor}
\usepackage{amsmath}
\usepackage{subfigure}
\usepackage{graphicx}
\usepackage{algorithm}
\usepackage{capt-of}

\newsavebox{\ieeealgbox}
\newenvironment{boxedalgorithmic}
  {\begin{lrbox}{\ieeealgbox}
   \begin{minipage}{\dimexpr\columnwidth-2\fboxsep-2\fboxrule}
   \begin{algorithmic}}
  {\end{algorithmic}
   \end{minipage}
   \end{lrbox}\noindent\fbox{\usebox{\ieeealgbox}}}
\makeatletter
\newcommand\fs@norules{\def\@fs@cfont{\bfseries}\let\@fs@capt\floatc@ruled
  \def\@fs@pre{}%
  \def\@fs@post{}%
  \def\@fs@mid{\kern3pt}%
  \let\@fs@iftopcapt\iftrue}
\makeatother
\floatstyle{norules}
\restylefloat{algorithm}
\newtheorem{definition}{Definition}[section]
\def\BibTeX{{\rm B\kern-.05em{\sc i\kern-.025em b}\kern-.08em
    T\kern-.1667em\lower.7ex\hbox{E}\kern-.125emX}}
\begin{document}


\title{Adversarial Profiles: Detecting Out-Distribution \& Adversarial Samples in Pre-trained CNNs
}
\author{\IEEEauthorblockN{Arezoo Rajabi}
\IEEEauthorblockA{\textit{School of EECS, Oregon State University } \\
rajabia@oregonstate.edu}
\and
\IEEEauthorblockN{Rakesh B. Bobba}
\IEEEauthorblockA{\textit{School of EECS, Oregon State University } \\
rakesh.bobba@oregonstate.edu}

}

\maketitle

\begin{abstract}
Despite high accuracy of Convolutional Neural Networks (CNNs), they are vulnerable to adversarial and out-distribution examples. There are many proposed methods that tend to detect or make CNNs robust against these fooling examples. However, most such methods need access to a wide range of fooling examples to retrain the network or to tune detection parameters. Here, we propose a method to detect adversarial and out-distribution examples against a \textit{pre-trained CNN} without needing to retrain the CNN or needing access to a wide variety of  fooling examples. To this end, we create \emph{adversarial profiles} for each class using only one adversarial attack generation technique. We then wrap a detector around the pre-trained CNN that applies the created adversarial profile to each input and uses the output to decide whether or not the input is legitimate. Our initial evaluation of this approach  using MNIST dataset show that \emph{adversarial profile} based detection is effective in detecting at least $92\%$ of out-distribution examples  and $59\%$ of adversarial examples.
\end{abstract}

\begin{IEEEkeywords}
CNN, Out-distribution, Adversarial examples
\end{IEEEkeywords}

\section{Introduction\label{sec:introduction}}

Convolutional Neural Networks(CNNs) have become popular due to their promising accuracy achievements on vision tasks such as image classification.  
Similar to many traditional classification methods, CNNs are typically trained only on data relevant to the task (in-distribution samples) and therefore they are only able to classify in-distribution samples correctly. In other words, CNNs assume that any received sample is from one of its in-distribution classes and classify it confidently~\cite{lee2017training,rajabi2018towards}. However, when such CNNs are deployed in real systems they may need to deal with images that do not belong to any of its in-distribution classes either inadvertently or because of adversarial intent. Hence, an adversary can fool CNNs with minimum effort. 

In addition to the vulnerability to out-distribution samples, it has also been shown that traditional CNNs are ill-equipped to deal with adversarial examples~\cite{goodfellow2014explaining}. Adversarial examples are in-distribution samples crafted by adding a learned noise (adversarial perturbation) to cause misclassification in the target CNN. In other words, an adversary can learn adversarial examples using his local CNN(s) to fool an unknown target CNN (black-box attacks)~\cite{goodfellow2014explaining}. An adversarial example conceptually belongs to the source image’s class, but for from the target CNN's perspective it belongs to another class. 

Many methods have been proposed for out-distribution and adversarial examples detection which need to train/retrain network(s) (e.g.,~\cite{lee2017training,xu2017feature,meng2017magnet,papernot2015distillation}) or  access or even train wide range of adversarial/out-distribution samples (to tune thresholds or add to training set)~\cite{lee2017training,rajabi2018towards,liang2017principled,madry2017towards}. Furthermore, most of proposed methods only address one of these issues not both (e.g., ~\cite{dathathri2018detecting, lee2017training}).
While, there are lot of high-accurate CNNs for different tasks and retraining all them to make robust against out-distribution/adversarial samples are computationally expensive. Moreover, users prefer to improve  available networks to make them capable of rejecting out-distribution samples.

Here, we propose to employ an adversarial attack itself in service of detecting out-distribution and adversarial  samples. Adversarial attack models generate adversarial examples that have  target class features  since they are misclassfied to the target class. Also, these examples have  the source class features because  they conceptually  belong to the source class. This property of adversarial examples can be utilized for learning  adversarial perturbations such that they result in misclassification of instances from a specific source class  to a specific target class (i.e., targeted adversarial examples). We propose to a create set of such targeted adversarial perturbations for each source class, where each perturbation is designed to move an instance of the source class to a different target class.  We refer to this set of adversarial perturbations as an \emph{adversarial profile} of the source class.

In this paper we propose to create and use \emph{adversarial profiles} to detect out-distributions and adversarial samples thrown at a pre-trained CNN without having to re-train the CNN in question or needing access to a wide variety of out-distribution and adversarial samples to train the detector. In particular we show that outputs of the CNN on instances of the input sample perturbed using adversarial profiles can be used to detect out-distribution samples and adversarial examples.  

We demonstrate that our proposed method can detect at least $92\%$ of out-distribution examples  and $59\%$ of adversarial examples for MNIST dataset. Although our proposed method has lower detection rate compared to previously proposed methods (e.g.~\cite{rajabi2018towards,xu2017feature}), unlike those methods our proposed method works for both out-distribution  and adversarial detection problems. Further, we only need a few in-distribution samples to learn adversarial profiles. Note that the adversarial profiles can be learned once (off-line) and be deployed in CNNs for out-distribution  and adversarial examples detection.

Briefly, our contributions are:
\begin{itemize}
    
    \item 	We aim to make existing CNNs robust against out-distribution examples without retraining them or requiring any out-distribution or adversarial samples for learning.
    \item We use an adversarial model to generate adversarial profiles for in-distribution classes to detect out-distribution samples.
    \item We evaluate our adversarial profiles against both out-distribution samples and black-box  adversarial examples.
\end{itemize}

\section{Preliminaries\label{sec: preliminaries }}
A CNN is a neural network with some convoulutional, ReLU and pooling layers followed by fully connected layers. A neural network is represented with $F(x)$ where $F$ is function that takes input images $x$ and returns a probability vector $[y_1, y_2,\cdots, y_c]$. The input $x$ is assigned to $i^{th}$ class if $y_i$ has the largest value. 

In~\cite{goodfellow2014explaining}, it was shown that CNNs are vulnerable to crafted images with small noise ($\delta$) learned over CNN. These carefully crafted images are called adversarial examples. Adversarial attacks can be targeted or untargeted. In a targeted attack, an adversary's aim is to make the given CNN misclassify the adversarial example to a target class ($j$):
\begin{equation}
\min \|\delta\|_p , \: \:\:s.t. \:\:\mathrm{argmax}\:\: F(x+\delta)= j, \:\: x+\delta \in [0,1]
\end{equation}

While in an untargeted attack, an adversary aims to make the target CNN misclassify adversarial example to any class ($\mathrm{argmax}\:\:F(x+\delta)\neq i$, where $i$ is the true label of input image $x$) other than the true class.  

CW~\cite{carlini2017towards} is one of the strongest known adversarial  attacks which we extend here  for generating adversarial profiles. 
Authors of~\cite{carlini2017towards} introduced a new objective function based on minimum difference between the probabilities assigned to the true class and target class such that the objective function has lower value if a CNN misclassifies the image with higher confidence as follows:
\begin{equation}
\begin{split}
f(x)=  \max(\max  \{F(x)_k: k\neq j\}-F(x)_{j} , -\kappa)
\end{split}
\end{equation}

Where $F(x)_k$ is the the probability  assigned to $k^{th}$ class by  CNN  for given  input ($x$). Also, $j$ and $\kappa$ denote the target class and confidence parameter, respectively.  
To ensure that the input values remain in range after adding the perturbation, in CW~\cite{carlini2017towards} 
the perturbation $\delta$ is added to  arctangent hyperbolic value of an image instead of added directly to images as follows:

\begin{small}
\begin{equation}\label{eq:carliniSum1}
\begin{split}
&z=\mathrm{arctanh} ((x-0.5)\times2) + \delta\\ &x_{pert.}= \frac{1}{2}(\tanh(z)+1)
\end{split}
\end{equation}
\end{small}

Where $x_{pert.}$ is the perturbed image $x$ and the variable $\delta$ can take any value from $[-\infty, \infty]$.

\begin{figure}[h]
    \centering
    \subfigure[Intra-class transferability\label{fig:ptable}]{\includegraphics[scale=0.28]{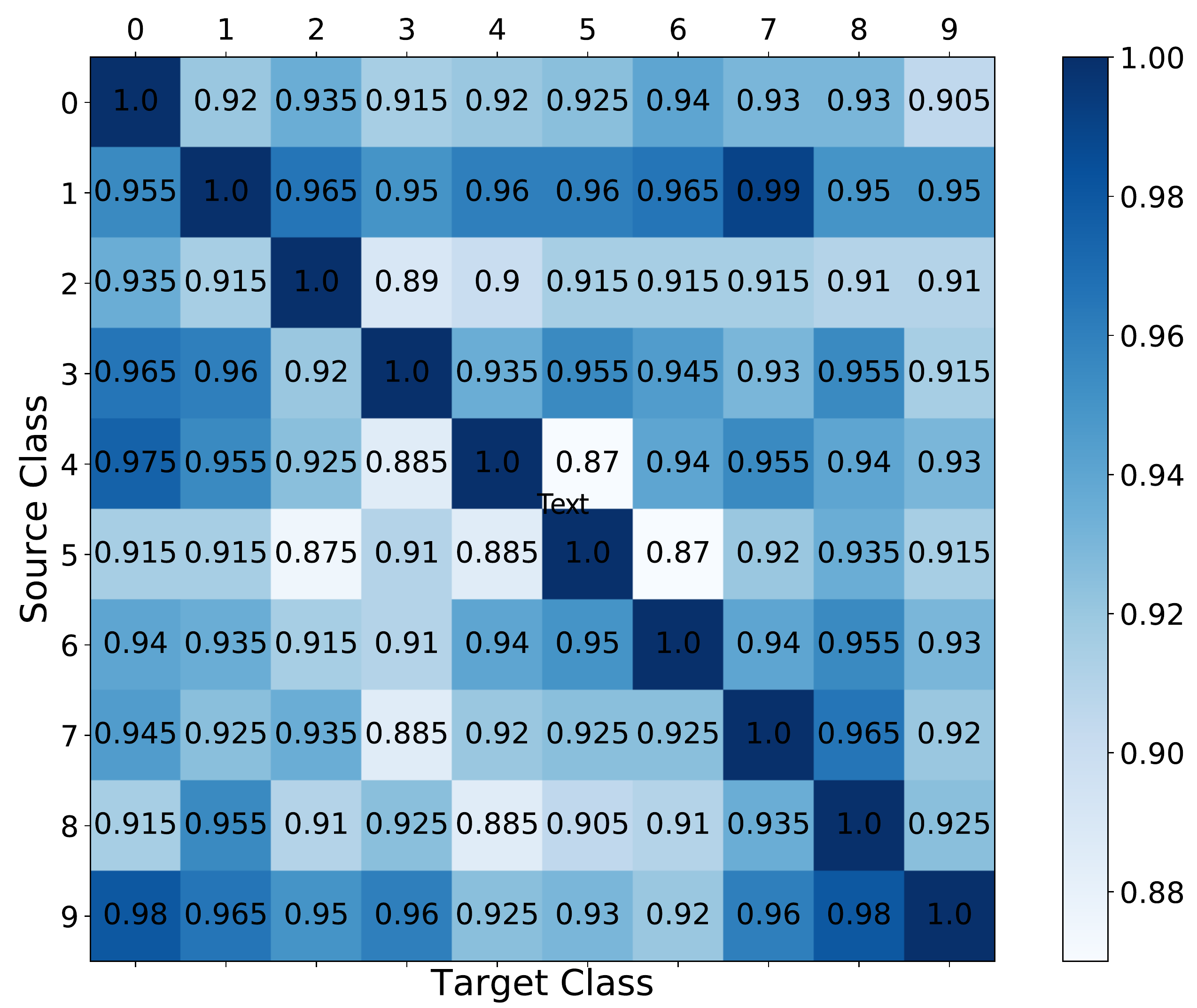}}
\subfigure[Inter-class transferability\label{fig:etable}]{\includegraphics[scale=0.28]{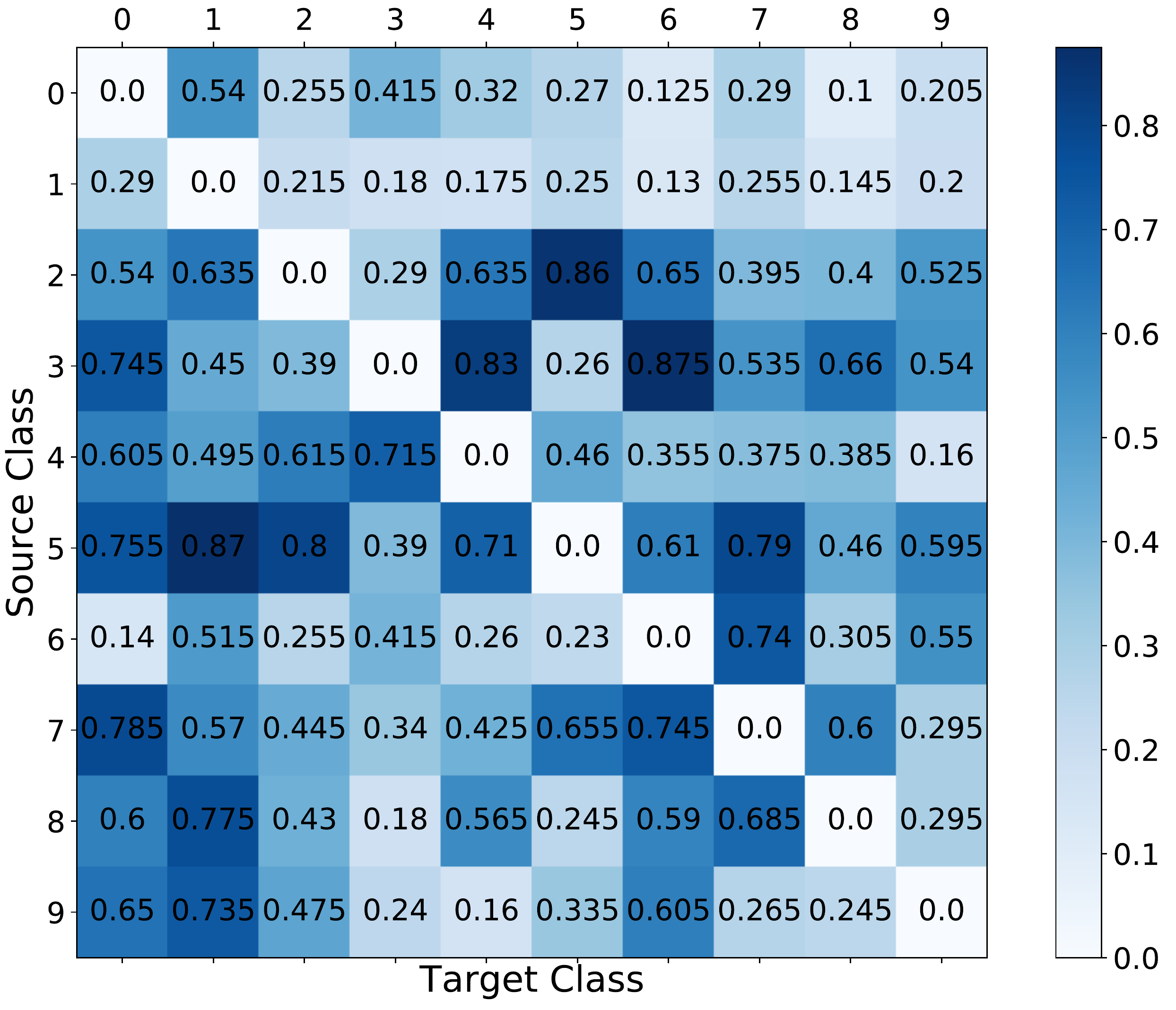}}
    \caption{Intra and Inter class transferability matrices.  The element at $[i,j]$  in Inter class transferability matrix represents the value of $p_{i,j}$. Similarly,  the element at $[i,j]$  in intra class transferability matrix  represents the value of $e_{i,j}$. The larger value for $p_{i,j}$ and lower value for $e_{i,j}$ are preferred.}
    \label{fig:PEtable}
\end{figure}
\section{Proposed Method}
In this section, we describe our proposed method for out-distribution and black-box adversarial examples detection. In our method, for each class we learn an adversarial profile that is specific that class. Let assume that the given in-distribution samples belong to $c$ classes.
\begin{definition}
For a given CNN, adversarial profile of $i^{th}$ class ($C_i$) is a set of adversarial perturbations $\{\delta_{i,1},\cdots, \delta_{i,i-1},  \delta_{i,i+1},\cdots, \delta_{i,c}\} $ that satisfy the following two properties: i) adding $\delta_{i,j}$  to any clean sample from class $i$ leads the target CNN to misclassify that sample to class $j$ (i.e., if $ x\in c_i , \:\:\: \mathrm{argmax}\:\:F(x+\delta_{i,j})=j$) with high probability; and ii) adding $\delta_{i,j}$ to any clean sample from other classes ($\neq i$), would lead the CNN to  misclassify that sample to any other class except $j$ (i.e., if $ x\notin c_i , \:\:\: \mathrm{argmax} \:\:F(x+\delta_{i,j})\neq j$).
\end{definition} with high probability.

In  the rest of this section, we first explain how we learn these adversarial profiles, and then demonstrate how to utilize them for detecting out-distribution  and adversarial examples. 

\noindent\textbf{Learning Adversarial Profile:}
Finding an adversarial perturbation $\delta_{i,j}$ that is be able to fool all samples from class $i$ to target class $j$ is hard and computationally expensive. Therefore,  we only use $n$ randomly selected samples from each source class to learn an adversarial perturbation and accept it for use in the adversarial profile if it can fool the CNN for at least $p*n$ of them ($0<p<1$). 

Increasing $p$ value increases the computational cost of learning adversarial profiles, but a large value of $p$ also leads to more powerful adversarial profile members (larger perturbation amount ($\delta_{ij}$)) that can fool  all samples from a given class to the respective target class more confidently. On the other hand, a low value of $p$ will result in adversarial profile whose members can not fool the source class samples to the target classes well. Therefore, selecting an appropriate $p$ value is very important.  

 We extended  CW~\cite{carlini2017towards} attack model to generate an adversarial perturbation that is able to work for a batch of images and get them misclassified to a targeted class as below.
 
\begin{small}
\begin{equation}
\begin{split}
  \delta_{i,j}=\min_{\delta} 
  &\sum_{x_t \in X} \max(\max_{k, k\neq j}  \{F(x_{pert.,t})_k\}-F(x_{pert.,t})_{j} , -\kappa)\\
\end{split}
\end{equation}
\end{small}
where $X$ is set of n inputs belonging to class $i$ ($\{x_t \in C_i|t=1\cdots n\}$) used for learning adversarial profiles.
\begin{algorithm}[htb]
\caption{\small Out-Distribution \& Adversarial Detection Detection~\label{alg:detection}}
\begin{boxedalgorithmic}[1]
\STATE $i\gets \mathrm{argmax}\:\: F(x)$
\STATE $\mathrm{score}\gets 0$
\FOR{$j \in \{1,\cdots, i-1,i+1,\cdots,C\}$}
\STATE  $\mathrm{score}\gets \mathrm{score}+ (p_{i,j}-e_{i,j})\times I(\mathrm{argmax}\:\: F(x)==j) $ 
\ENDFOR
\STATE $\%\mathrm{I(True)}=1 \:\: \& \:\:  \mathrm{I(False)}=0$
\STATE $\mathrm{score}\gets \frac{\mathrm{score}}{\sum_{j}(p_{i,j}-e_{i,j}))}$
\STATE $\%$ Select K random profiles out of $|C|-1$
\STATE $ S\gets \mathrm{random-select}(K,C-\{i\})$ 
\FOR{$k \in S $ }
\STATE $\mathrm{score}_k\gets 0$
\FOR {$j \in \{1,\cdots, k-1,k+1,\cdots,C\}$}
\IF {$\mathrm{score}>\tau$}
\STATE $\mathrm{score}_k\gets \mathrm{score}_k+ (p_{k,j}-e_{k,j})\times I(\mathrm{argmax}\:\:F(x)\neq j) $ 
\ELSE
\STATE $\mathrm{score}_k\gets \mathrm{score}_k+ (p_{k,j}-e_{k,j})\times I(\mathrm{argmax}\:\:F(x)= j) $
\ENDIF
\ENDFOR
\ENDFOR
\STATE  $\mathrm{score}\gets \mathrm{score}+\sum_k \frac{\mathrm{score}_k}{\sum_j (p_{k,j}-e_{k,j})}$
\IF {$\frac{\mathrm{score}}{K+1}>\tau$} 
\STATE return  $i$ 
\ELSE 
\STATE \textbf{Null}
\ENDIF
\end{boxedalgorithmic}
\end{algorithm}

\noindent\textbf{Detection Algorithm:} To detect out-distribution and adversarial examples, we evaluate the behaviour of input data after applying an adversarial profile. All perturbations in an adversarial profile do not have same effect. To assess a perturbation's effectiveness we define two metrics,  \emph{Intra-Class Transferability} and \emph{Inter-Class Transferability}.

\begin{definition}
We say an adversarial perturbation $\delta_{i,j}$ is \emph{$p_{i,j}$-intra-class transferable} if the probability of fooling CNN to the target class $j$ for samples from source class $i$ is $p_{i,j}$ (i.e., $p(\mathrm{argmax}\:\: F(x+\delta_{i,j})==j|x\in C_i)=p_{i,j}$).
\end{definition}

Although, the adversarial perturbation $\delta_{i,j}$ is learned on source class $i$ but there is a possibility that this perturbation also works on samples from other classes ($\neq i$). To measure transferability of an adversarial perturbation for other classes we define  \emph{Inter-Class Transferability}. 

\begin{definition}
We call adversarial perturbation $\delta_{i,j}$ is \emph{$e_{i,j}$-inter-class-transferable} if the probability of fooling the  CNN to the target class $j$ for samples from other classes ($\neq i$) is $e_{i,j}$ (i.e., $p(\mathrm{argmax}\:\: F(x+\delta_{i,j})==j|x\notin C_i)=e_{i,j}$).
\end{definition}

To estimate intra-class transferability ($p_{i,j}$) of a perturbation $\delta_{i,j}$, we sample  from source class $i$ and apply the perturbation $\delta_{i,j}$ on the samples and measure how many of  them are misclassified to the target class $j$. To estimate inter-class transferability ($e_{i,j}$), we sample  from other classes ($ \neq i $) and apply adversarial profile $\delta_{i,j}$ on them. Then we measure how many of  them are misclssified to the target class $j$.  An ideal adversarial profile member $\delta_{i,j}$  has a large value (close to $1$) of inter-class transferability ($p_{i,j}$) and  a small value (close to $0$) of intra-Class transferability ($e_{i,j}$). We give preference to ideal perturbations by using $p_{i,j} - e_{i,j}$ as weights in the detection decision process (see line 4 in Algorithm ~\ref{alg:detection}).  

Note that out-distribution samples do not belong to in-distribution space and therefore they do not have source or target class features. Hence applying an adversarial perturbation designed for in-distribution samples to out-distribution samples should induce a different behavior than expected. To detect out-distribution samples, we apply the perturbations from the adversarial profile associated with the assigned class  on the input sample and compute a score that measures how close its behaviour is to that expected of an in-distribution sample of the assigned class.  The score we use is as follows:

\begin{equation}
    \mathrm{score}=\frac{1}{\sum_{j} (p_{i,j}-e_{i,j}) }\sum_{j} (p_{i,j}-e_{i,j}) \times I(\mathrm{argmax}\:\: F(x)==j)
\end{equation}

For a clean in-distribution sample, and an ideal adversarial profile this score will be close to $1$ as all perturbations will take the sample to the expected target class. For out-distribution samples, with high probability this will be less than $1$. We find and use a detection threshold to distinguish between in- and out-distributions samples (see Section~\ref{sec:eval}). 


Unlike out-distribution samples that may not have any in-distribution class features, adversarial samples tend to have features of both original class (albeit weakened) and fooling class. Hence, when an adversarial profile associated with the fooling class is applied it tends to work well making it hard to distinguish from clean samples. However, unlike with clean class samples, for adversarial samples adversarial profile associated with other classes also tends to work well because of weakened features due to the perturbation. We leverage this difference to detect adversarial samples (see line 14 in Algorithm ~\ref{alg:detection}). As shown in Algorithm~\ref{alg:detection}, to increase the accuracy, we test every input sample with the adversarial profile of the assigned class, and also using adversarial profiles of $K$ other randomly selected classes.



\noindent{\bf Usage Example:} Let us assume $x$ is a sample submitted to CNN for classification. Let us say that the CNN classifies this samples to class $i$. If $x$ is indeed an instance of class $i$, then perturbations in  adversarial profile associated with $C_i$ when applied to $x$ will lead the CNN to classify the perturbed samples to their respective target classes with high probability. On the other hand, perturbing the sample using members from adversarial profiles associated with other classes $C_j$ ($j \neq i$) will \emph{not} take the sample to their target class with high probability.

However, if $x$ is an out-distribution sample then then perturbations in  adversarial profile associated with $C_i$ when applied to $x$ will not \emph{consistently} lead the CNN to classify the perturbed samples to their respective target classes. On the other hand if $x$ is an adversarial sample, then adversarial profiles associated with both class $C_i$ and other classes $C_j$ ($j \neq i$) can potentially take the sample to their target class. 


\section{Evaluation}\label{sec:eval}

In this section we evaluate our proposed method on MNIST~\cite{lecun1998mnist} dataset. This dataset contains $60000$ images of hand-writing  digits ($\{0,1,\cdots,,9\}$), $50k$ for training and $10k$ images for testing in each class.  The learned CNN reached to $99.2\%$ accuracy on test data. We sampled  $100$ images for each class from test data to create adversarial profiles with $p$ value of $0.9$ which means adversarial profiles can successfully fool at least $90\%$ of samples used for training  to the target class. To assess learned adversarial examples, we randomly took $n=100$ samples from each class  and applied adversarial profile  on them. we estimate the intra-class transferability ($p_{i,j}$) and inter-class transferability ($e_{i,j}$) as following:
\begin{equation}
\begin{split}
    p_{i,j}&=\frac{\sum_i^n I(\mathrm{argmax}\:\:F(x+\delta_{i,j})==j)}{n}\\
    e_{i,j}&=\frac{\sum_i^n I(\mathrm{argmax}\:\:F(x+\delta_{i,j})==j)}{n}\\
\end{split}
\end{equation}
 The estimated $p_{i,j}$ and $e_{i,j}$ values for MNIST dataset are shown in Figure~\ref{fig:PEtable}. For example, adversarial perturbation $\delta_{5,4}$ has intra-class transferability value of $0.885$ which means that this adversarial profile can cause $88.5\%$ of the samples of class $5$ to be misclassified to class $4$. Moreover this adversarial perturbation causes samples from other classes to misclassify to target class $4$ with probability of $71\%$. Therefore, this adversarial perturbation is not ideal and will have less weight ($0.885-0.71 = 0.175$) in our decision making process. On the other hand $\delta_{0,8}$ is a good perturbation because it has a high intra-class transferability of $93\%$ and a low inter-class transferability of $10\%$ and will be weighted more ($0.93-0.10 = 0.83$).  
\begin{table}[h]
 \caption{\small Adversarial and out-distribution detection rates. \label{tab:out-detection}}
    \centering
    \begin{tabular}{|c|c|c|}
    \hline
         Input & Detection Rate & Parameters \\ \hline
         SVHN & 97\%& - \\ \hline
         ImageNet & 92\% & - \\ \hline
         CIFAR-100 & 92\% & - \\ \hline
         LSUN & 97\% & -\\ \hline
         FGS~\cite{goodfellow2014explaining}& 63\%& $\epsilon=0.2$ \\ \hline
         T-FGS~\cite{goodfellow2014explaining}& 65\%& $\epsilon=0.2$ \\ \hline
         DeepFool~\cite{moosavi2016deepfool}& 59\%& \\ \hline
         I-FGS~\cite{kurakin2017adversarial}& 64\%& $\epsilon=0.02$, $\alpha=0.2$ \# of iter. $=20$ \\ \hline
         
    \end{tabular}
\end{table}
To estimate the best threshold $\tau$, we minimized false positive rate and maximized detection rate on  in-distribution samples as follow:
\begin{equation}
\begin{split}
  \tau&= \max_{\tau} \sum_{i} \frac{1}{2}[ I(\mathrm{score}(x_i+\delta_{i,j}|x_i\in C_i)>\tau) +\\ &I(\mathrm{score}(x_i+\delta_{i,j}|x_i\notin C_i)<\tau)] 
\end{split}
\end{equation}

To estimate $\mathrm{score}(x_i+\delta_{i,j}|x_i\in C_i)>\tau $, we took $100$ samples from different classes  and then applied out-distribution method detection with their true label. Similarly, to estimate and $\mathrm{score}(x_i+\delta_{i,j}|x_i\notin C_i)<\tau$, we used the same samples with  labels different from true labels, therefore adversarial profiles of the fake labels will be applied on them (to satisfy condition of $x_i\notin C_i$).

We considered the $4$ datasets, SVHN~\cite{netzer2011reading}, ImageNet~\cite{krizhevsky2012imagenet}, LSUN~\cite{yu15lsun} and CIFAR-100~\cite{krizhevsky2009learning} as out-distribution samples to evaluate our method. We sampled $100$ images from each dataset and applied our detection algorithm on them.
As shown in Table~\ref{tab:out-detection}, our proposed method is able to reject at least $92\%$ of out-distribution samples from ImageNet and CIFAR-100, and as high as $97\%$ from SVHN and LSUN. Similar to out-distribution samples, we used $4$ different attack models (FGS,T-FGS,DeepFool,I-FGS) to generate $n=100$ adversarial test examples. As shown in Table~\ref{tab:out-detection}, our method can detect between $59\% - 65\%$ of adversarial examples (DeepFool and T-FGS respectively). Compared to previously proposed methods (e.g.~\cite{rajabi2018towards,xu2017feature}), our approach has a lower detection rate. But in contrast to previous work, we do not need to retrain the CNN, and we do not need to have access to a lot of out-distribution or adversarial samples. The only thing we need is a small in-distribution set to make a given CNN robust against out-distribution samples and majority of adversarial samples.
\section{Conclusion}
In this paper, we proposed using adversarial profiles to detect out-distribution and adversarial examples  without retraining the model or having to train on examples  of  out-distribution or adversarial examples. We deploy an adversarial generation model to create a set of profiles for each class. These profiles are adversarial perturbations that can lead samples of the source class to misclassify to target classes. We show that we can learn an adversarial profile for each class with high intra-class transferability and low inter-class transferability.  Our evaluation on MNIST dataset shows that our proposed method can detect at least $92\%$ of out-distribution examples and $59\%$ of adversarial examples. 

In the future we would like to evaluate this approach on other datasets (e.g., CIFAR). Further we would like to explore more sophisticated detectors instead of using simple threshold based detection. 
\bibliographystyle{IEEEtran}  
\bibliography{MyBib}  
\end{document}